\pdfoutput=1
\documentclass[sigconf]{acmart}

\AtBeginDocument{%
  }

\copyrightyear{2026}
\acmYear{2026}
\setcopyright{cc}
\setcctype{by-nc-nd}
\acmConference[SA Technical Communications '26]{SIGGRAPH Asia 2026 Technical Communications}{December 01--04, 2026}{Kuala Lumpur, Malaysia}
\acmBooktitle{SIGGRAPH Asia 2026 Technical Communications (SA Technical Communications '26), December 01--04, 2026, Kuala Lumpur, Malaysia}
\acmDOI{10.1145/3829339.3847830}
\acmISBN{979-8-4007-2841-9/2026/12}

\usepackage{subcaption}
\usepackage{xcolor}
\usepackage{graphicx}
\usepackage{tikz}
\usetikzlibrary{arrows.meta}

\newcommand{\NVERTS}{8{,}447}
\newcommand{\NTRAIN}{347}
\newcommand{\NTEST}{59}
\newcommand{\NEXPR}{19}

\begin{document}

\title{FACET: A Data-Efficient Regional Parametric Head Model}

\author{Mathieu Lamarre}
\email{mlamarre@ea.com}
\orcid{0009-0008-3965-1173}
\affiliation{%
  \institution{Electronic Arts}
  \city{Montr\'eal}
  \country{Canada}
}

\author{Elly Akhoundi}
\email{eakhoundi@ea.com}
\orcid{0009-0006-4989-1731}
\affiliation{%
  \institution{Electronic Arts}
  \city{Vancouver}
  \country{Canada}
}

\author{\mbox{Thibault Goyallon}}
\email{tgoyallon@ea.com}
\orcid{0000-0003-3238-3707}
\affiliation{%
  \institution{Electronic Arts}
  \city{Toronto}
  \country{Canada}
}

\author{Patrick Anderson}
\email{panderson@ea.com}
\orcid{0009-0004-4386-4789}
\affiliation{%
  \institution{Electronic Arts}
  \city{Montr\'eal}
  \country{Canada}
}

\author{\'Etienne Danvoye}
\email{edanvoye@ea.com}
\orcid{0009-0006-4837-9541}
\affiliation{%
  \institution{Electronic Arts}
  \city{Montr\'eal}
  \country{Canada}
}

\renewcommand{\shortauthors}{Lamarre et al.}

\begin{abstract}
Production face pipelines need parametric head models that are expressive, FACS-controllable, and buildable on a small, fixed pool of identities.
More training identities would always help, but high-quality registered head scans are expensive to capture; when scaling the data is not an option, the remaining lever is architecture.
We present FACET, a parametric head model that makes the most of a fixed identity pool through architectural locality rather than more scans, and makes three contributions, each isolated against a matched global-PCA ablation on our own topology, training set, and normalization.
(1) A regional identity decomposition into 16 anatomical regions bypasses the $N{-}1$ rank ceiling of a global PCA, driving held-out neutral-fit error from a $1.95$~mm floor down to $0.23$~mm on the same training identities.
(2) A learned FACS action-unit expression model with per-identity neural personalization not only stays FACS-controllable but \emph{beats} a properly-regularized global expression PCA at every component count---refuting the argument that FACS is too overcomplete to fit.
(3) A dedicated neck/shoulder region reduces pose leakage into the pose-invariant upper face during unposing.
Together, these show that architectural locality is the effective lever for a production head model under a fixed identity budget.

\end{abstract}



\begin{teaserfigure}%
\centering%
\begin{tikzpicture}[x=1cm,y=1cm]%
  \node[anchor=west] (p1) at (0,0)%
    {\includegraphics[width=0.298\linewidth]{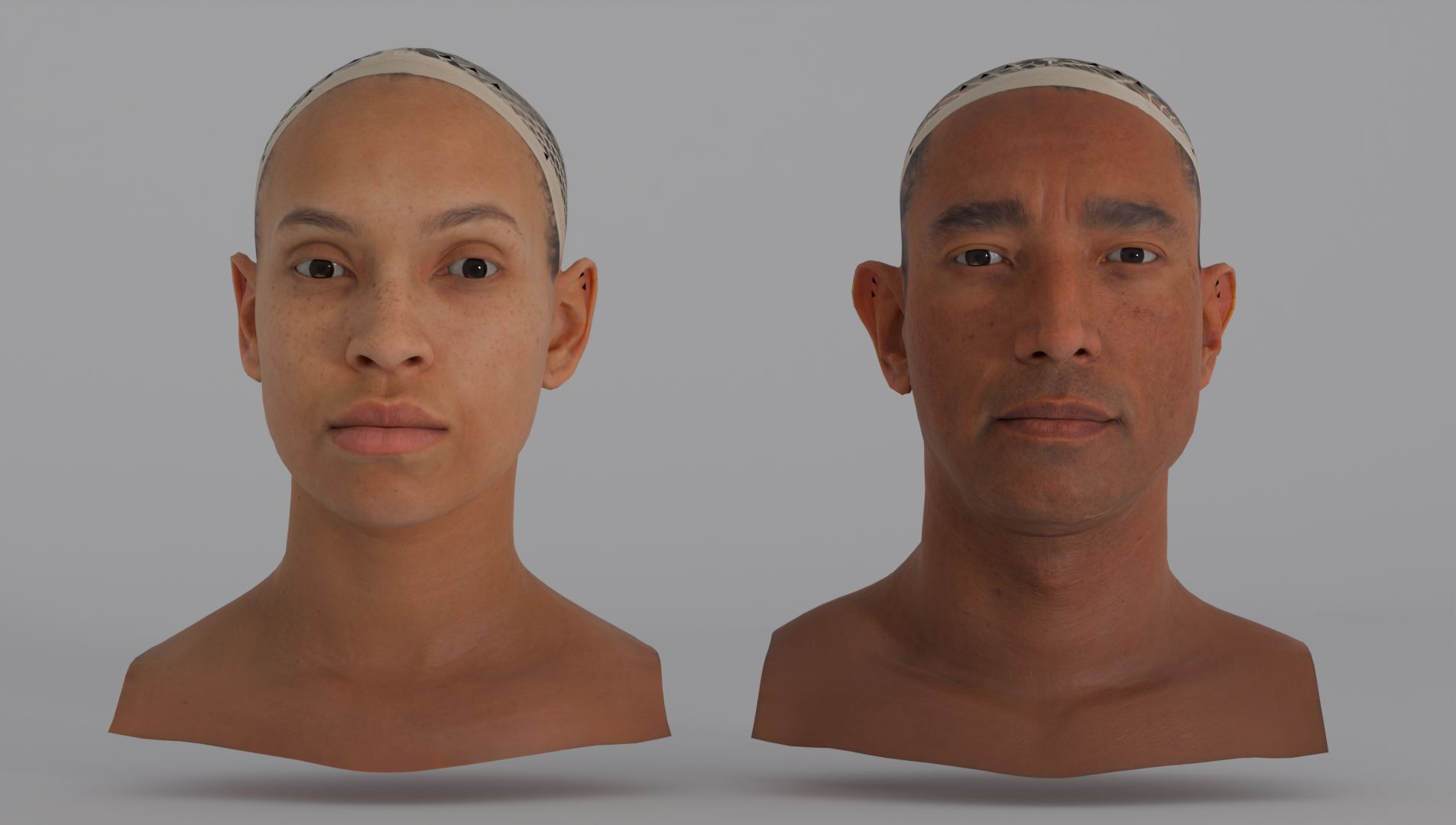}};%
  \node[anchor=west] (p2) at (0.345\linewidth,0)%
    {\includegraphics[width=0.298\linewidth]{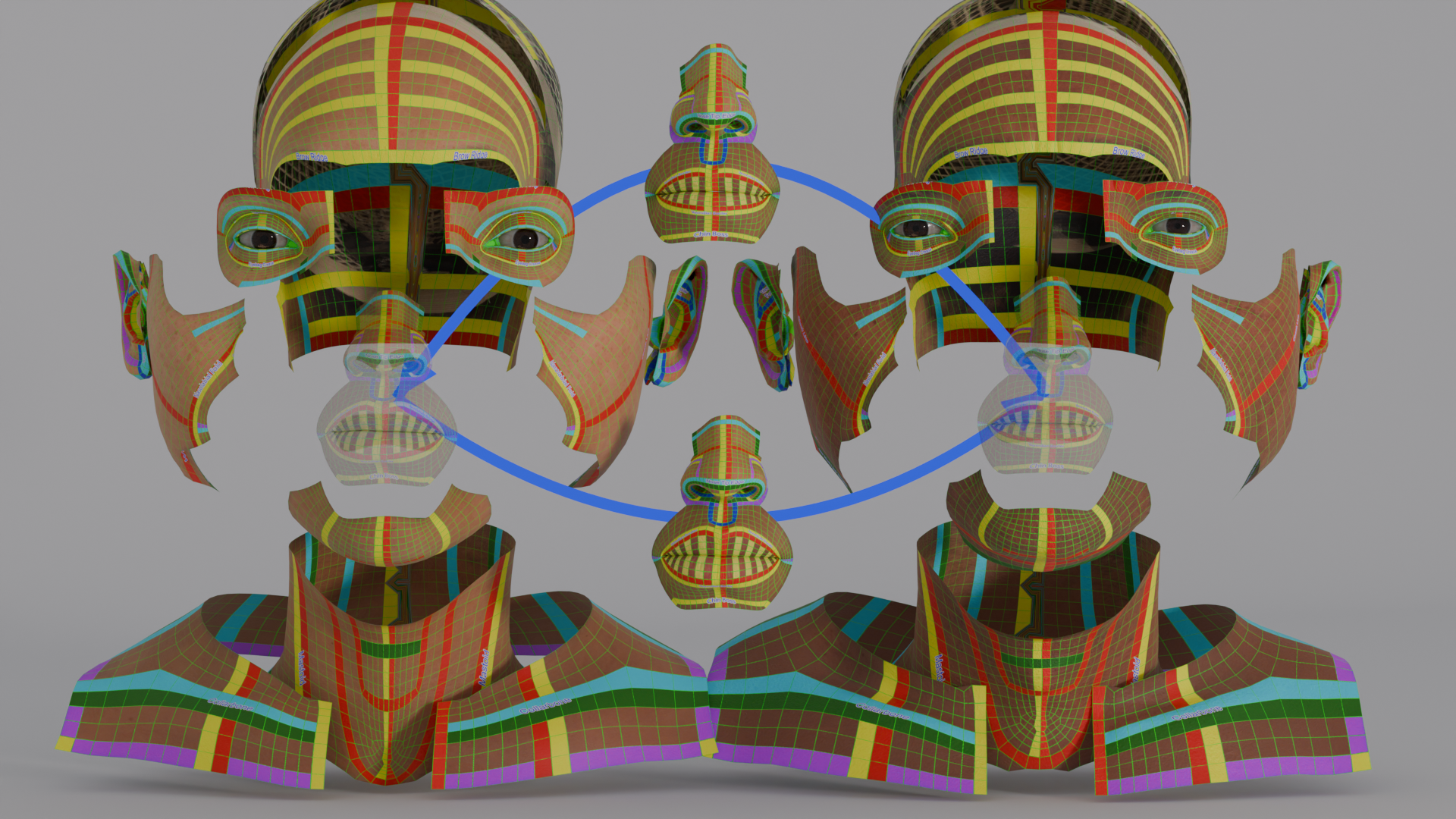}};%
  \node[anchor=west] (p3) at (0.690\linewidth,0)%
    {\includegraphics[width=0.298\linewidth]{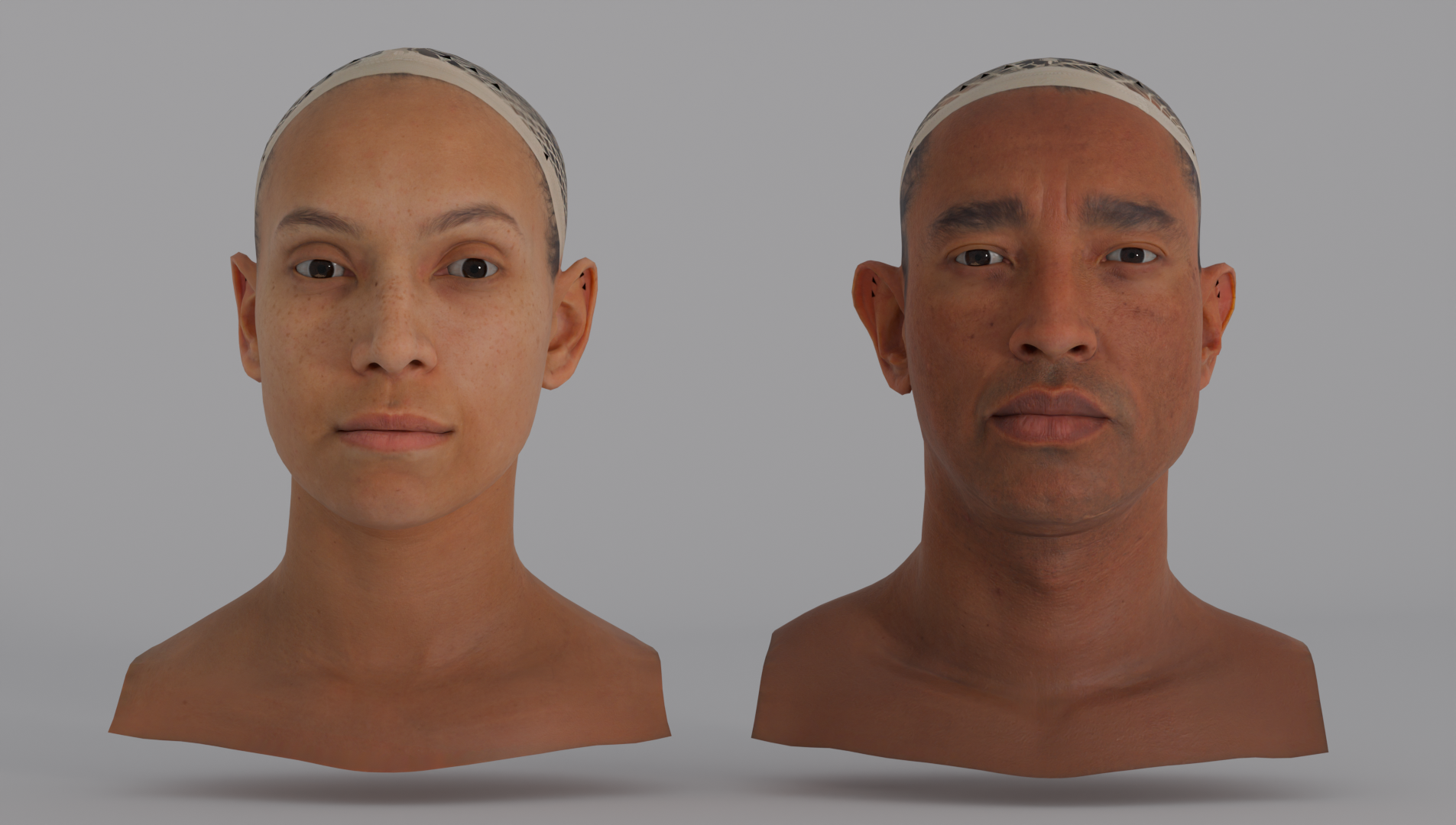}};%
  \draw[-{Latex[length=2.4mm]},line width=0.5mm] (p1.east) -- (p2.west);%
  \draw[-{Latex[length=2.4mm]},line width=0.5mm] (p2.east) -- (p3.west);%
\end{tikzpicture}%
\caption{FACET represents identity per-region across 16 anatomical regions (coloured
segmentation, center; mouth interior not shown), so features transfer locally.
\textbf{Left:} two input identities. \textbf{Center:} only the nose and lips
are swapped, snapped mid-crossing. \textbf{Right:} each head reconstructs
seamlessly with the other's nose and lips---an edit a global PCA model cannot
make without perturbing the whole face.}%
\Description{Three rendered panels. Left: two input head identities side by side. Center:
each head's nose and lip regions arc across to the opposite head along curved
trails, snapped mid-crossing so the two sets appear one above the other, with
fading copies indicating motion. Right: the two heads reconstructed with their
nose and lips swapped.}%
\label{fig:teaser}%
\end{teaserfigure}

\maketitle

\section{Introduction}
\label{sec:introduction}

Parametric head models underpin production facial pipelines: a compact, riggable, DCC-compatible representation an artist can fit, edit, and animate.
The dominant recipe, exemplified by FLAME~\cite{li2017flame}, learns global PCA identity and expression bases from a large scan corpus, and increasing fidelity has largely meant acquiring more scans---LSFM~\cite{booth2018lsfm} reaches nearly ten thousand identities.
For a production studio, though, high-quality registered head scans are expensive to capture, so the identity pool is a fixed set of a few hundred subjects rather than the thousands used by large-scale models.
The question is therefore how to extract the most from a small pool.

When the identity pool is fixed, a global identity PCA on $N$ subjects spans at most $N{-}1$ modes and no scan quality lifts that ceiling.
The remaining lever is architecture: how much a model extracts from the identities it has.
We substantiate this across the three axes of a complete head model with FACET.
First, FACET decomposes identity into 16 anatomical regions, each with its own PCA basis blended by Laplacian handles; because each region is supported by the same subjects but spans a smaller patch, it exposes more usable identity dimensions than a basis capped at $N{-}1$ and improves held-out generalization---data efficiency, not merely compact editing.
Second, it drives expression with learned FACS action-unit blendshapes personalized per identity by a small neural network, staying FACS-controllable while reusing an existing FACS-annotated capture deck.
Third, it adds a dedicated neck/shoulder region that confines articulation during unposing, reducing pose leakage into the pose-invariant upper face.

\section{Related Work}
\label{sec:related}

\paragraph{Identity and regional shape models.}
Linear 3D morphable models capture identity variation with a single global PCA basis, and the dominant strategy for increasing fidelity has been to acquire more scans: FLAME~\cite{li2017flame} learns its shape space from thousands of head scans, and LSFM~\cite{booth2018lsfm} scales this to nearly ten thousand identities.
A global basis trained on $N$ identities is, however, hard-capped at $N-1$ modes, which is restrictive when the training pool is a small, fixed set of subjects.
Localized models predate this concern: Tena et al.~\cite{tena2011regionpca} split the face into region-based linear sub-models that generalize to unseen data better than a holistic basis, and NPHM~\cite{giebenhain2023nphm} localizes an implicit head into per-region MLPs.
These works motivate regionality through animation control, editing, or representational power.
FACET instead frames regional PCA over 16 anatomical regions as a \emph{data-efficiency} mechanism that bypasses the $N-1$ rank ceiling, and measures the resulting improvement in held-out generalization on a small identity pool.

\paragraph{Expression: FACS versus statistical bases.}
FLAME~\cite{li2017flame} explicitly rejects a FACS parameterization as overcomplete and adopts an orthonormal global expression PCA, an argument that shaped much subsequent work.
FACS-driven bases nonetheless remain the lingua franca of production: ICT-FaceKit~\cite{li2020ictfacekit} pairs hand-crafted action-unit blendshapes with a global identity PCA, and FaceScape~\cite{yang2020facescape} derives per-subject FACS blendshapes from a bilinear model fit to twenty captured expressions per subject.
None of these establish whether a semantic FACS basis can match a global PCA basis in fitting fidelity under matched conditions.
FACET supplies that controlled comparison: a learned FACS-AU expression model with per-identity neural personalization (hash-grid and FiLM correctives) is shown to \emph{exceed} a FLAME-class global expression PCA at every component count on the same mesh and training set, while remaining FACS-controllable.

\section{Method}
\label{sec:method}

\begin{figure}[t]
\centering
\includegraphics[width=\linewidth,keepaspectratio]{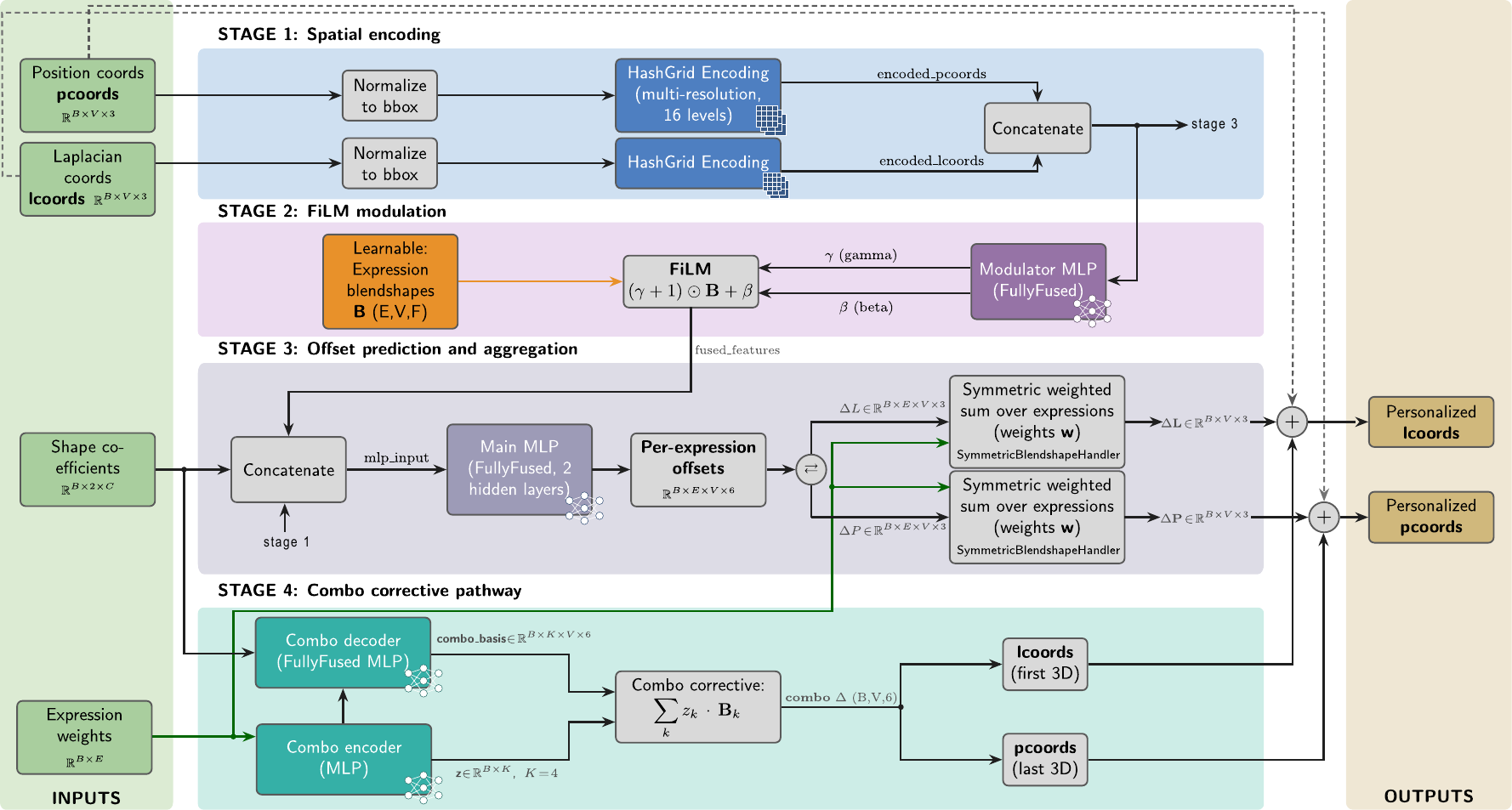}
\caption{FACET's per-identity expression personalization network.}
\Description{Block diagram of the personalization network: position and
Laplacian hash-grid encodings feed a FiLM modulator over learnable action-unit
blendshapes, a main MLP, a symmetric weighted sum, and a parallel combo encoder
and geometry-conditioned decoder that meet in a final linear combination.}
\label{fig:personalization}
\end{figure}

FACET is a production parametric head model on a single quad-dominant topology of \NVERTS{} vertices, comprising three components: a regional identity model, a FACS action-unit expression model with per-identity neural personalization, and a head\slash neck\slash shoulder articulation with unposing.

\subsection{Regional identity model}
\label{sec:method-identity}

A global identity PCA trained on $N$ registered neutral scans spans at most $N{-}1$ modes.
On a fixed identity pool, this rank ceiling---not scan resolution---bounds representable identity detail.
FACET instead partitions the head into 16 anatomical regions and fits separate position- and Laplacian-coordinate PCA bases per region.
Because each region draws its modes from the same $N$ identities but spans a lower-dimensional patch of the surface, each region contributes up to $2(N{-}1)$ data-supported coefficients.
For the \NTRAIN{}-identity training pool used here, the regional ceiling is therefore $16\cdot2\cdot(\NTRAIN{}{-}1)=11{,}072$ coefficients, rather than $N{-}1=346$ for one global PCA---data efficiency, not merely compact editing.

Unlike a global model, which forms vertices by adding shape offsets to a template, FACET reassembles the regions with a sparse linear solve.
Each region's PCA predicts target position and Laplacian coordinates; the head vertices $\mathbf{p}$ are recovered as the weighted least-squares fit
\begin{equation}
\begin{split}
\mathbf{p}^\star(\boldsymbol{\beta}) &= \operatorname*{arg\,min}_{\mathbf{p}}\; \bigl\lVert \mathbf{W}_L\,(\mathbf{L}\mathbf{p} - \boldsymbol{\delta}) \bigr\rVert^2 + \lambda^2 \bigl\lVert \mathbf{W}_P\,(\mathbf{p} - \boldsymbol{\xi}) \bigr\rVert^2, \\
\boldsymbol{\delta}(\boldsymbol{\beta}^L) &= \boldsymbol{\mu}_L + \sum_{r=1}^{16} \mathbf{D}^L_r\,\boldsymbol{\beta}^L_r,\quad
\boldsymbol{\xi}(\boldsymbol{\beta}^P) = \boldsymbol{\mu}_P + \sum_{r=1}^{16} \mathbf{D}^P_r\,\boldsymbol{\beta}^P_r,
\end{split}
\label{eq:regional-identity}
\end{equation}
where $\mathbf{L}$ is the cotangent Laplacian; $\mathbf{D}^L_r,\boldsymbol{\beta}^L_r$ and $\mathbf{D}^P_r,\boldsymbol{\beta}^P_r$ are the Laplacian- and position-coordinate PCA bases and coefficients for region~$r$; and $\lambda$ is the main soft-position weight, fixed at $0.58$ for all reported results.
Per-vertex weights $\mathbf{W}_L$ and $\mathbf{W}_P$ overlap across boundaries.
Because the overlapping weights are built into the system rather than applied after the fact, the assembled head is seamless and remains a single editable mesh~\cite{botsch2008deformation}.
FACET factors the normal-equation system once at construction and reuses the Large Steps GPU Cholesky solver~\cite{nicolet2021large} for subsequent fits.
Locality also makes features transferable: an artist can swap one subject's nose or lips onto another without perturbing the rest of the face (Fig.~\ref{fig:teaser}).

\subsection{FACS expression model with neural personalization}
\label{sec:method-expression}

Rather than a global expression PCA, FACET drives expression with a bank of learnable blendshapes indexed by FACS action units, keeping the model semantically controllable and compatible with an existing FACS-annotated capture deck.
A generic AU rig cannot capture how the same action unit deforms different faces, so we personalize the blendshapes per identity with a small neural network per region (Fig.~\ref{fig:personalization}).
Two hash-grid encodings---on mesh position and on Laplacian coordinates---form a spatial code that FiLM-modulates the learnable AU blendshape tensor; a compact main MLP (one hidden layer of 16 neurons) then consumes the modulated blendshapes, the spatial code, and the identity's shape coefficients to predict per-expression offsets in a joint 6D (position$+$\allowbreak Laplacian) output, which a symmetric weighted sum over the AU weights aggregates into an expression displacement.
Each region stores only its left and center action units, mirroring them to the right by reflection or splitting a midline unit by vertex masking---roughly halving its blendshapes.
The hash encoders and MLPs use the fully-fused Instant-NGP~\cite{mueller2022instant} implementation, keeping personalization practical at interactive fitting rates.
Formally, the expression offset at a vertex $\mathbf{x}$ is
\begin{equation}
\begin{split}
\mathbf{E}(\mathbf{w}, \boldsymbol{\beta}) = {}& \sum_{a=1}^{A} w_a\, \mathcal{N}_\theta\!\bigl(\phi(\mathbf{x}),\; (\boldsymbol{\gamma}{+}\mathbf{1}) \odot \mathbf{B}_a + \boldsymbol{\tau},\; \boldsymbol{\beta}\bigr) \\
& + \sum_{k=1}^{4} g_k(\mathbf{w})\, \mathbf{C}_k(\mathbf{x}; \boldsymbol{\beta}),
\end{split}
\label{eq:facs-expression}
\end{equation}
where $w_a$ activates FACS action unit $a$, $\mathbf{B}_a$ is its learnable blendshape, $\phi(\mathbf{x})$ the concatenated position/Laplacian hash-grid code, $(\boldsymbol{\gamma},\boldsymbol{\tau})$ the FiLM parameters regressed from $\phi(\mathbf{x})$, $\mathcal{N}_\theta$ the main MLP conditioned on the fitted shape coefficients $\boldsymbol{\beta}$, and $g_k(\mathbf{w})$ the combo encoder driving a geometry-conditioned basis $\mathbf{C}_k$.
Like the identity terms, these 6D offsets enter the solve of Eq.~\ref{eq:regional-identity} as additional Laplacian and position targets.
Replacing $\mathcal{N}_\theta$ with a fixed, identity-independent basis and the semantic weights $w_a$ with abstract coefficients recovers a global expression PCA in the style of FLAME~\cite{li2017flame}.
The combo term handles AU-combination correctives: a small encoder maps AU weights to $K{=}4$ coefficients, while the geometry-conditioned decoder (independent of the AU weights) emits a per-identity combo basis; the two meet only in a final linear combination.
That independence makes the combo basis \emph{bakeable}---at deployment the decoder runs once per identity and freezes into a static table, so per-frame fitting touches only the tiny AU-weight encoder.
The rig thus keeps FACS semantics and real-time evaluation while adapting every action unit to the fitted identity.

We bootstrap personalization with a deformation-transfer prior: each neutral is a per-region blend of eight production rigs, whose expressions transfer onto the identity to supply AU-labeled data and resolve identifiability.
The same transfer over our cleaned expression corpus fills gaps so every identity has the same expressions, and training alternates per-sample pose/AU fits with global personalization passes.

\subsection{Articulation and unposing}
\label{sec:method-unposing}

FACET articulates jaw, neck, and eyes with linear blend skinning~\cite{li2017flame} and adds a dedicated neck/shoulder region.
Without such a split, the shoulder and neck motion in every posed 4D scan has nowhere to go and leaks into the identity geometry during unposing; FACET's regions confine it, leaving the pose-invariant upper face nearly unchanged.
Section~\ref{sec:results} evaluates this by how much the upper face moves.

\section{Results}
\label{sec:results}

\paragraph{Shared protocol.}
All variants use the FACET topology, an identical eye-placement normalization, and the same training pool of \NTRAIN{} consented identities; every weight in any compared model is estimated from this pool only.
We evaluate on disjoint held-out test sets: \NTEST{} distinct neutral scans for identity (E1), and the same \NTEST{} identities across \NEXPR{} expressions each for expression (E2).
We report per-vertex vertex-to-vertex (V2V) error in mm on the held-out sets.
For each claim we compare against a \emph{matched} global-PCA ablation---not the released FLAME model---that shares this setup and differs only in the architectural variable under test.

\begin{figure}[t]
\centering
\includegraphics[width=\linewidth]{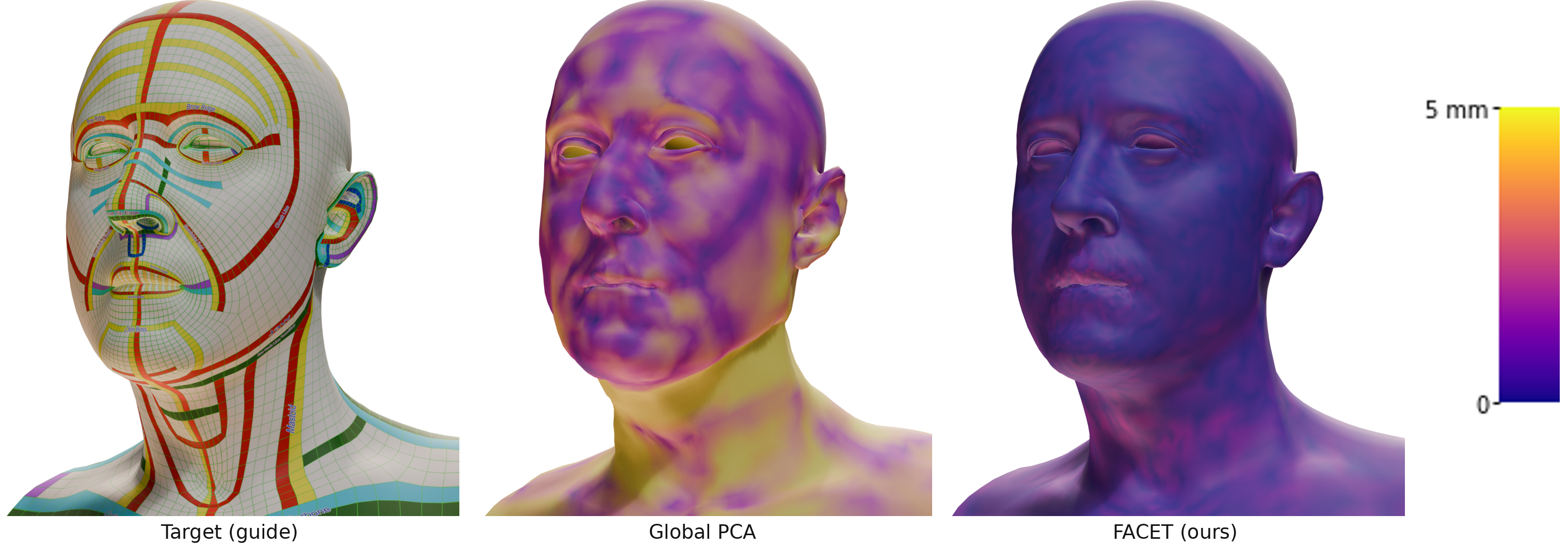}
\caption{Held-out identity fit for one subject, per-vertex error on a shared
0--5~mm scale. Left: target with guide texture. Middle: global PCA is warm across
neck, jaw and cheeks. Right: regional FACET stays cold nearly everywhere.}
\Description{Three head renders in a row on a shared color scale: a guide-textured
target, a global PCA render with warm error over the neck and lower face, and a
FACET render that is almost uniformly cold purple.}
\label{fig:e1-pseudocolor}
\end{figure}

\begin{figure}[t]
\centering
\includegraphics[width=0.98\linewidth]{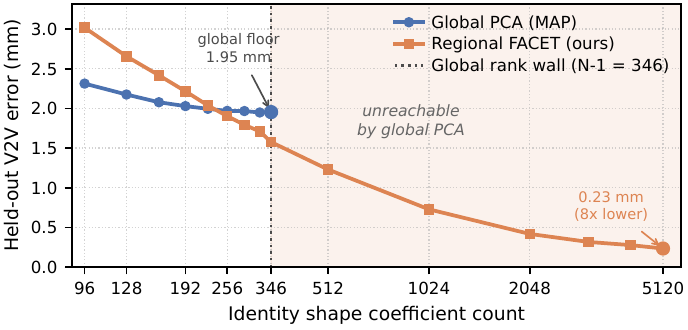}
\\[2pt]
\includegraphics[width=0.98\linewidth]{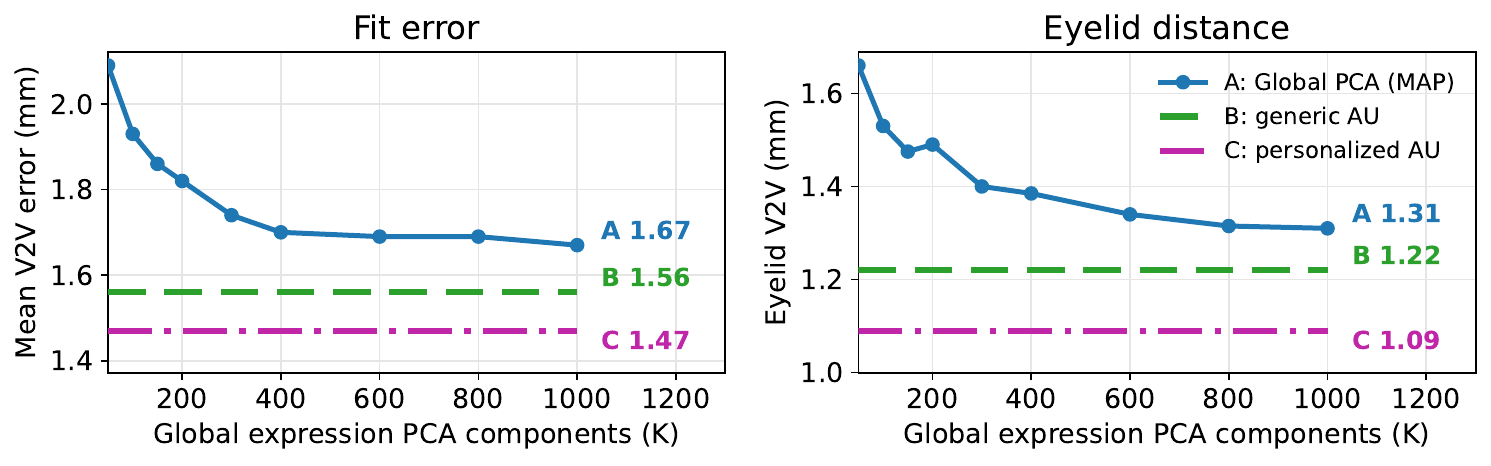}
\caption{Held-out fitting error against coefficient count; values are labelled
in the plots. \emph{Top (E1):} a global identity PCA floors at the
rank-$(N{-}1)=346$ wall while regional FACET keeps descending past it.
\emph{Bottom (E2):} the generic (B) and personalized (C) FACS-AU variants sit
below the global expression PCA (A) at every $K$, in fit error (left) and eyelid
V2V (right).}
\Description{Two line charts. The top identity chart shows a global PCA curve
ending at a vertical rank wall while a regional FACET curve continues far below
it. The bottom pair of expression charts shows the global PCA curve staying above
two horizontal reference lines for the generic and personalized action-unit
models.}
\label{fig:datacurve}
\end{figure}

\paragraph{E1: regional decomposition is data-efficient.}
Figure~\ref{fig:datacurve} plots held-out neutral-fit V2V error against the total identity coefficient count.
The global PCA baseline descends but terminates at the rank-\allowbreak$(N{-}1)=346$ wall, where it floors at $1.95$~mm: no additional coefficients exist on a fixed pool of \NTRAIN{} identities.
FACET's regional model crosses under the global floor near $K\approx320$ and keeps descending into the regime the global basis structurally cannot reach, down to $0.23$~mm mean (median $0.17$~mm, p95 $0.60$~mm) at $K{=}5120$---roughly $8.3\times$ lower than the global floor, on the \emph{same} training identities.
Because the improvement is measured on held-out scans, the extra dimensions improve generalization rather than overfit.
The gains concentrate in the regions where locality matters most---nose, eye orbits, lips (Table~\ref{tab:main})---as the per-vertex error map for one held-out subject makes visible (Fig.~\ref{fig:e1-pseudocolor}).

\paragraph{E2: personalized FACS expression dominates a global PCA basis.}
FLAME~\cite{li2017flame} rejected FACS as overcomplete and hard to fit in favor of a global expression PCA.
We test this on the fitted-identity protocol: each neutral is frozen with production FACET, then only expression parameters are optimized.
The global PCA variant (A) uses a standard eigenvalue-weighted MAP coefficient prior $\lambda\sum_i (c_i/\sigma_i)^2$~\cite{blanz1999morphable} with $\lambda{=}5\!\times\!10^{-5}$ held fixed across all $K$; the eigenvalue weighting makes this scale-invariant to $K$.
This is a stronger prior than FLAME's plain squared-norm, so the comparison does not understate global PCA's fitting capacity.
Figure~\ref{fig:datacurve} (bottom) sweeps variant A from $K{=}50$ to $K{=}1000$ against the generic (B) and personalized (C) AU variants.
Both AU variants beat global PCA at \emph{every} $K$, including $K{=}1000$ where PCA has the most capacity: mean fit error settles at $1.67$~mm (A) versus $1.56$~mm (B) and $1.47$~mm (C), with $C<B<A$ throughout.
The eyelid metric---PCA's hardest case, since eye closure is where a global basis struggles most---is sharpest: $1.31$~mm (A) versus $1.22$~mm (B) and $1.09$~mm (C).
The AU model is uniformly better while remaining FACS-controllable.

\begin{table}[t]
\centering
\caption{Held-out V2V error (mm). E1 at full model capacity; E2 at $K{=}1000$.}
\label{tab:main}
\small
\begin{tabular}{@{}lcc@{}}
\toprule
\multicolumn{3}{@{}l}{\textbf{E1 --- Identity (neutral fit), V2V mm}} \\
Region & Global PCA & Regional FACET \\
\midrule
Overall    & 1.95 & \textbf{0.23} \\
Nose       & 1.23 & \textbf{0.12} \\
Eye orbit  & 1.52 & \textbf{0.13} \\
Lips       & 1.46 & \textbf{0.15} \\
Ear        & 1.99 & \textbf{0.27} \\
\midrule
\multicolumn{3}{@{}l}{\textbf{E2 --- Expression fit ($K{=}1000$), V2V mm}} \\
 & Global PCA (A) & Pers. AU (C) \\
\midrule
Fit error  & 1.67 & \textbf{1.47} \\
Eyelid     & 1.31 & \textbf{1.09} \\
\bottomrule
\end{tabular}
\end{table}

\begin{figure}[t]
\input{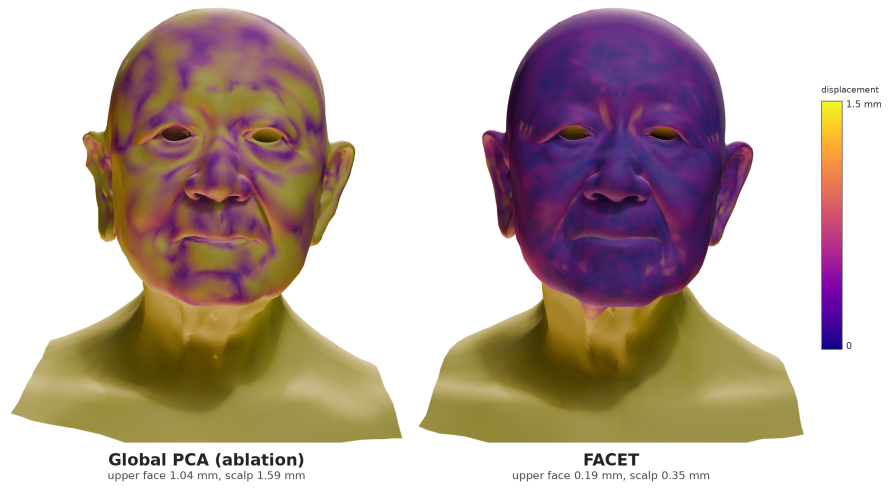}
\caption{Per-vertex displacement (mm) between each model's unposed canonical
head and the observed posed scan. Without a dedicated neck/shoulder region
(left) a global-PCA model leaks articulation into the pose-invariant upper face
and crown; FACET (right) confines it.}
\Description{Two pseudo-colored head renders with a shared displacement colorbar:
the global-PCA ablation panel has hot colors on the upper face and crown, while
the FACET panel remains mostly dark in those regions.}
\label{fig:unposing}
\end{figure}

\paragraph{E3: neck/shoulder region reduces pose leakage.}
The pose-invariant upper face (forehead and eye orbits) should not move when a model unposes neck and shoulder articulation.
A global-PCA head model has no localized neck/shoulder split, so fitting it to posed neutral scans leaks pose into the upper-face identity geometry.
Across the \NTRAIN{}-identity pool, mean upper-face V2V drops from $0.50$~mm (global-PCA ablation) to $0.16$~mm (production FACET), a $\sim\!3.1\times$ reduction.
Whole-head displacement is comparable ($1.54$ vs $1.58$~mm mean), so the gap is specifically pose leaking into the upper face (Fig.~\ref{fig:unposing}).

\paragraph{Ethics and likeness.}
All captures were collected from participants under informed-consent agreements.

\section{Discussion and Conclusion}
\label{sec:discussion}

When the identity pool is fixed, architectural locality is the lever: regional decomposition escapes the global rank ceiling, personalized FACS AUs beat global expression PCA while staying controllable, and a neck/shoulder region blocks pose leakage into the upper face.

\paragraph{Limitations.}
Our evaluation uses small held-out pools (\NTEST{} identities; \NTEST{}$\times$\NEXPR{} expressions), reflecting the fixed identity pool that motivates the work.

\paragraph{Conclusion.}
FACET shows that when more scans are out of reach, a production head model on a fixed identity pool is well served by architectural locality---regional identity, personalized FACS expression, and localized articulation---with each choice validated by a matched ablation that isolates it.

\bibliographystyle{ACM-Reference-Format}
\bibliography{facet}

\end{document}